\documentclass{article}

\usepackage{arxiv}
\usepackage[utf8]{inputenc}
\usepackage[T1]{fontenc}
\usepackage{amsmath}
\usepackage{amssymb}
\usepackage{amsfonts}
\usepackage{booktabs}
\usepackage{float}
\usepackage{graphicx}
\usepackage{microtype}
\usepackage[numbers]{natbib}
\usepackage{placeins}
\usepackage{tabularx}
\usepackage{url}
\usepackage{hyperref}
\usepackage{doi}

\newcolumntype{C}{>{\centering\arraybackslash}X}

\title{Accurate Recognition of Pneumonia and COVID-19 by Geometric Shape Normalization of Lung Region using Automatic Landmark Detection and Piecewise Affine Warping}

\author{
Salvador E. Ayala-Raggi\thanks{Correspondence: \texttt{salvador.raggi@correo.buap.mx}} \\
Facultad de Ciencias de la Electr\'onica\\
Benem\'erita Universidad Aut\'onoma de Puebla\\
Puebla, M\'exico\\
\texttt{salvador.raggi@correo.buap.mx}
\And
Rafael Alejandro Cruz-Ovando\\
Facultad de Ciencias de la Electr\'onica\\
Benem\'erita Universidad Aut\'onoma de Puebla\\
Puebla, M\'exico\\
\texttt{rafaelcruz.ovando@gmail.com}
\And
Lauro Reyes-Cocoletzi\\
Facultad de Ciencias de la Electr\'onica\\
Benem\'erita Universidad Aut\'onoma de Puebla\\
Puebla, M\'exico\\
\texttt{lreyesc@ipn.mx}
\And
Aldrin Barreto-Flores\\
Facultad de Ciencias de la Electr\'onica\\
Benem\'erita Universidad Aut\'onoma de Puebla\\
Puebla, M\'exico\\
\texttt{aldrin.barreto@correo.buap.mx}
}

\date{}

\renewcommand{\headeright}{arXiv Preprint}
\renewcommand{\undertitle}{arXiv Preprint}
\renewcommand{\shorttitle}{This manuscript is currently under consideration for publication.}

\hypersetup{
pdftitle={Accurate Recognition of Pneumonia and COVID-19 by Geometric Shape Normalization of Lung Region using Automatic Landmark Detection and Piecewise Affine Warping},
pdfsubject={cs.CV},
pdfauthor={Salvador E. Ayala-Raggi, Rafael Alejandro Cruz-Ovando, Lauro Reyes-Cocoletzi, Aldrin Barreto-Flores},
pdfkeywords={COVID-19, pneumonia recognition, landmark detection, piecewise affine warping, chest X-ray, lung region segmentation}
}

\begin{document}
\maketitle

\begin{abstract}
This paper presents an automatic classification system for pulmonary diseases in chest X-rays based on geometric normalization. The proposed method consists of three main modules. \textbf{Module 1: A landmark detector:} A ResNet-18 convolutional neural network with coordinate attention mechanism is trained to predict 15 landmarks defining the lung contour, achieving a mean error of 3.61 pixels (median 3.07 pixels) through an ensemble of four models with test-time augmentation. \textbf{Module 2: Geometric normalizer:} a set of landmarks surrounding the lung region is used to geometrically normalize each image. This normalization involves: Generalized Procrustes Analysis used once to obtain a standard lung shape, Delaunay triangulation to build a deformation mesh, and a piecewise affine transformation (warping) to map the original lung region to a standardized region. This process eliminates variations in position, scale, and orientation in the original set. \textbf{Module 3: Classifier:} normalized images are classified into three categories (COVID-19, Viral Pneumonia, and Normal) using a ResNet-18 classifier with transfer learning and a contrast adjustment (using SAHS method). The classifier was evaluated through five-fold cross-validation on the COVID-19 Radiography Database, demonstrating high stability with 98.60$\pm$0.26\% accuracy, 98.00\% F1-Macro, confirming the robustness of the approach. Although the classifier trained with original images reached a higher accuracy than using normalized images, Grad-CAM analysis and the cropping experiment suggest that this advantage is partly driven by acquisition artifacts rather than lung pathology. In contrast, geometrically normalized images outperform their non-aligned artifact-masked/cropped counterparts: 98.60\% vs.\ 96.24\% on the COVID-19 Radiography Database and 94.67\% vs.\ 94.17\% on a balanced adult--pediatric mixed dataset including pediatric cases from the Kermany dataset, suggesting that anatomical alignment can yield a more reliable and artifact-resistant representation for pulmonary disease recognition.
\end{abstract}

\keywords{lungs \and pulmonary diseases \and COVID-19 \and pneumonia \and diagnosis of pulmonary diseases \and chest X-ray image analysis \and automatic image analysis \and lung region segmentation \and landmark detection in medical images \and piecewise affine warping.}

\section{Introduction}

Chest X-rays constitute a first-line diagnostic tool for pneumonia due to their wide availability, low cost, and rapid acquisition~\cite{who2020chest}. However, manual interpretation requires specialized experience and is subject to inter-observer variability, motivating the development of automatic classification methods.

Deep learning-based methods have demonstrated great potential for automatic pneumonia detection~\cite{rajpurkar2017chexnet,wang2020covidnet,chowdhury2020can}. However, their performance can be affected by variations in image acquisition, including differences in patient position, scale, and orientation, as well as artifacts from each radiographic equipment. This phenomenon, known as \textit{domain shift}~\cite{zech2018variable}, can compromise the generalization capability of models.

\subsection{Related Work}

Various works have addressed variability in medical images through normalization and alignment techniques. Ayala-Raggi et al.~\cite{ayala2023synergizing} demonstrated in their study presented at ACPR 2023 that chest X-ray normalization, combined with discriminant feature selection, significantly improves automatic COVID-19 recognition by eliminating non-pathological variations. Picazo-Castillo et al.~\cite{picazo2024comparative} showed that the way visual information is presented to the classifier has a significant impact on performance, highlighting the importance of adequate preprocessing. Other approaches have explored the use of geometric transformations in medical images and include end-to-end learnable transformations such as Spatial Transformer Networks~\cite{jaderberg2015spatial}, though methods based on anatomical landmarks offer a more interpretable alternative with lower data requirements.

\subsection{Proposal of the Present Work}

We propose a geometric normalization method that goes beyond traditional rigid transformations (rotation, translation, scaling) by incorporating local deformations based on lung anatomy. The method uses three components: (1) convolutional neural networks to predict 15 landmarks defining the bilateral lung contour, (2) Generalized Procrustes Analysis~\cite{gower1975generalized} on manually annotated landmarks to determine a standard lung shape, and (3) Delaunay triangulation~\cite{delaunay1934sphere} with piecewise affine warping~\cite{wolberg1990digital} to deform each image to the standard shape. The normalized images are classified using a ResNet-18~\cite{he2016deep} with transfer learning.

This approach minimizes non-pathological variability, allowing the classifier to focus on intrinsic pathological characteristics. To the best of our knowledge, this method for COVID-19 classification is among the first to apply piecewise affine warping based on anatomical landmarks automatically detected, combining classical image processing techniques with modern deep learning in a fully automatic and interpretable pipeline.

\section{Materials and Methods}

\subsection{System Overview}

\begin{figure}[!htbp]
\centering
\includegraphics[width=\textwidth]{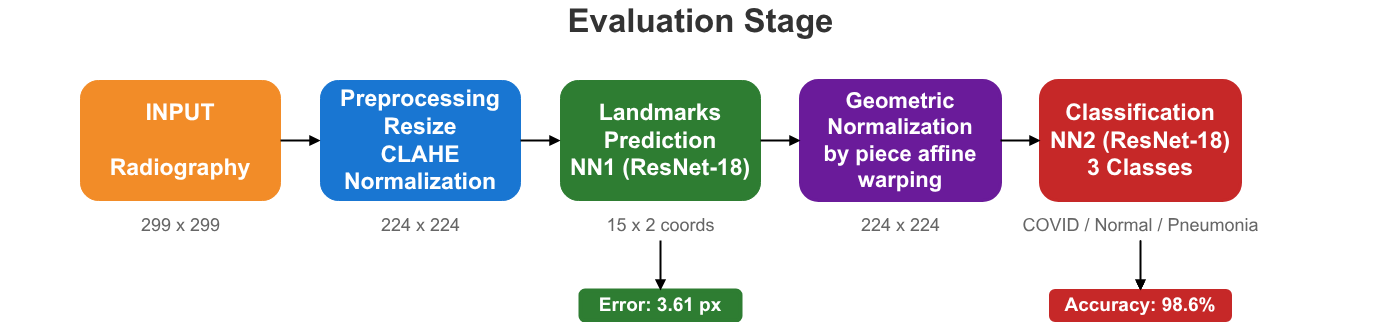}
\caption{Evaluation pipeline: input radiograph (299$\times$299) $\rightarrow$ preprocessing (224$\times$224, CLAHE) $\rightarrow$ landmark prediction (15 coordinate pairs, 3.61 px error) $\rightarrow$ piecewise affine warping $\rightarrow$ classification (COVID-19, Normal, Viral Pneumonia).}
\label{fig:flujo}
\end{figure}

The proposed system receives a chest X-ray and produces a classification output (Normal, COVID-19, or Viral Pneumonia) through three stages: (1) landmark prediction, (2) geometric normalization, and (3) classification. This modular architecture allows each component to be trained and evaluated independently. Figure~\ref{fig:flujo} illustrates the complete processing flow at evaluation.

\subsection{Dataset}

The public \textbf{COVID-19 Radiography Database}~\cite{chowdhury2020can,rahman2021exploring} was used, containing 15,153 posteroanterior radiographs organized into three categories: COVID-19 (3,616 images), Normal (10,192 images), and Viral Pneumonia (1,345 images). Original images (299$\times$299 pixels) were resized to 224$\times$224 pixels, standard format for ImageNet-pretrained models. For supervised training of the landmark prediction model, 15 landmarks were manually annotated by a single individual on the lung contour of 957 images. The annotation process required approximately three minutes per image (about 48 hours of cumulative annotation effort). Landmarks define control points on the bilateral lung silhouette, organized in a central vertical axis (5 points) and five symmetric left/right pairs (10 points). The dataset was split in a stratified manner into training (75\%), validation (15\%), and test (10\%).

\subsection{Landmark Prediction Model Architecture}

\begin{figure}[!htbp]
\centering
\includegraphics[width=0.95\textwidth]{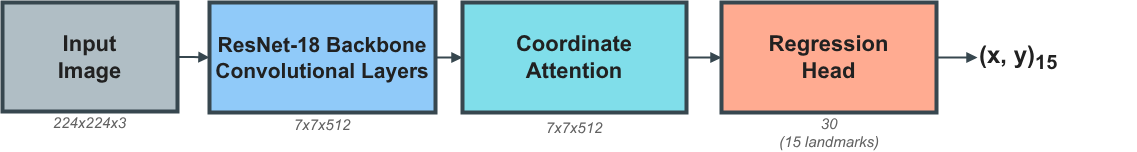}
\caption{Landmark prediction model architecture ($NN$ $1$). The input image passes through the ResNet-18 backbone (7$\times$7$\times$512 feature map), then through the Coordinate Attention module, and finally through the regression head producing 30 coordinate values.}
\label{fig:arquitectura}
\end{figure}

The landmark prediction model is based on a ResNet-18~\cite{he2016deep} architecture, referred to as $NN$ $1$, pretrained on ImageNet~\cite{deng2009imagenet}. The architecture comprises three components (Figure~\ref{fig:arquitectura}):

\begin{itemize}
\item \textbf{Backbone:} ResNet-18 convolutional layers process the input image (224$\times$224$\times$3) producing a 7$\times$7$\times$512 feature map.
\item \textbf{Coordinate Attention Module:} Inserted between the backbone and regression head, this mechanism~\cite{hou2021coordinate} captures positional dependencies along both spatial dimensions, which are crucial for precise localization tasks.
\item \textbf{Regression Head:} Three fully connected layers (512$\rightarrow$512$\rightarrow$768$\rightarrow$30) with Group Normalization~\cite{wu2018group} and staggered dropout. The output is 30 normalized values in [0,1], corresponding to the $(x,y)$ coordinates of the 15 landmarks.
\end{itemize}

\subsection{Landmark Model Training}

Wing Loss~\cite{feng2018wing} was used as the loss function, specifically designed for landmark localization due to its adaptive behavior: logarithmic for small errors (fine refinement) and linear for large errors (stability):

\begin{equation}
\mathcal{L}_{\mathrm{wing}}(e) =
\begin{cases}
\omega \ln\left(1 + \frac{|e|}{\epsilon}\right), & |e| < \omega \\
|e| - C, & |e| \geq \omega
\end{cases}
\end{equation}

with parameters $\omega = 10$ pixels, $\epsilon = 2$ pixels, and $C = \omega - \omega \ln(1 + \omega/\epsilon)$.

Training was organized in two phases. In the first phase (15 epochs), only the regression head was trained with a learning rate $1 \times 10^{-3}$, keeping the backbone and Coordinate Attention frozen. In the second phase (100 epochs maximum), all layers were trained with differentiated rates: $2 \times 10^{-5}$ for backbone and Coordinate Attention, and $2 \times 10^{-4}$ for the head.

To improve precision and reduce variance, an \textbf{ensemble} of four models trained with different random seeds (123, 321, 111, 666) was implemented. During inference, \textbf{Test-Time Augmentation (TTA)} was applied through horizontal flipping with symmetric landmark correction, averaging predictions from original and reflected images.

\subsection{Generalized Procrustes Analysis}

To determine a standard lung shape for normalization, Generalized Procrustes Analysis (GPA)~\cite{gower1975generalized} was applied to the 957 manually annotated landmark configurations.

\begin{figure}[!htbp]
\centering
\includegraphics[width=0.85\textwidth]{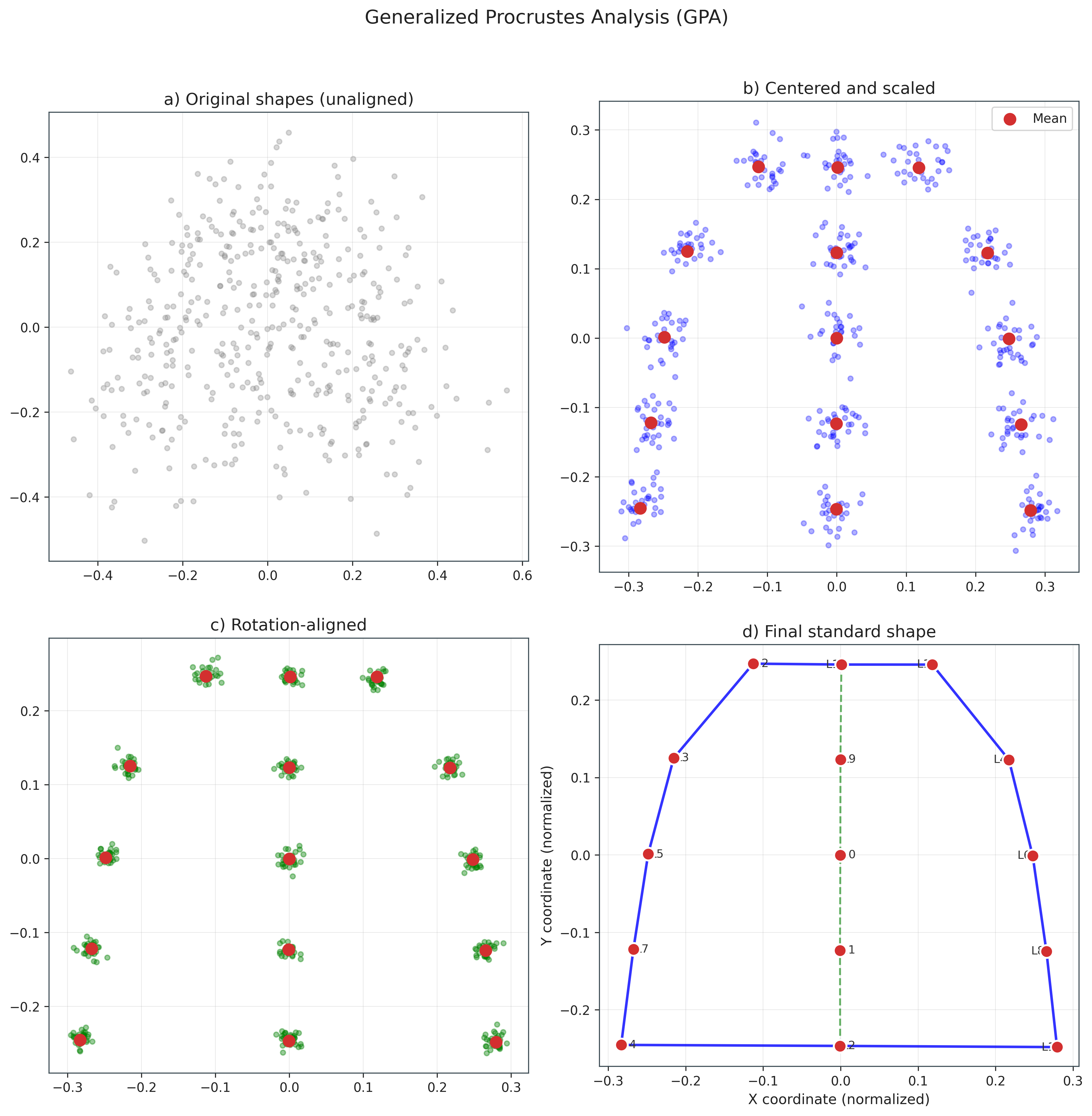}
\caption{GPA process: (a) original unaligned shapes, (b) centered and scaled configurations, (c) rotation-aligned configurations, (d) final standard shape with 15 labeled landmarks (L1--L15).}
\label{fig:gpa}
\end{figure}

GPA is an iterative algorithm that eliminates translation, scale, and rotation variations through three operations per configuration: (1) centering to origin, (2) scaling to unit norm, and (3) optimal rotation calculated by Singular Value Decomposition (SVD). The process repeats until convergence (tolerance $\tau = 10^{-8}$), producing a standard shape that represents the average lung morphology of the dataset (Figure~\ref{fig:gpa}).

\subsection{Delaunay Triangulation and Warping}

\begin{figure}[!htbp]
\centering
\includegraphics[width=0.55\textwidth]{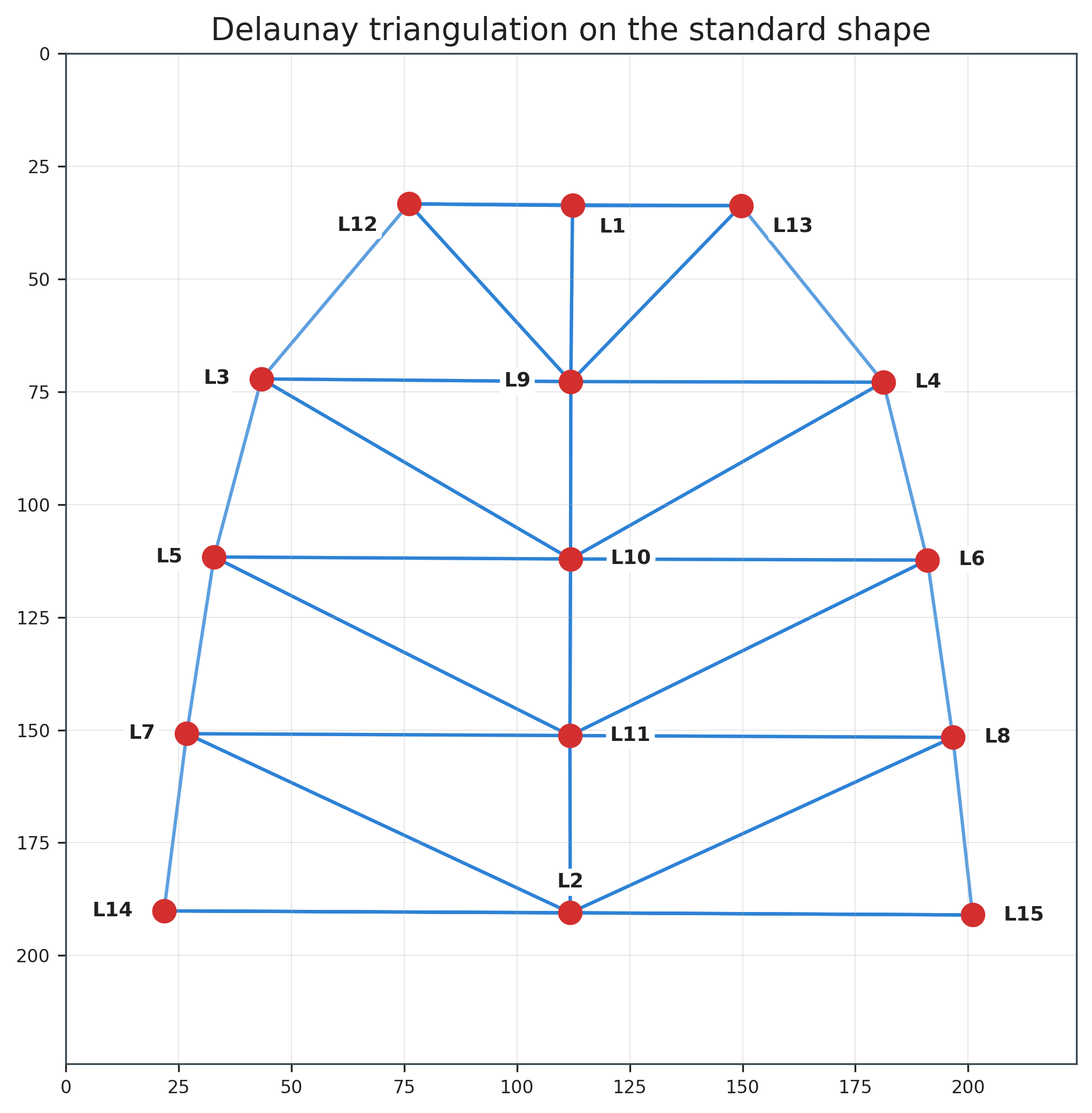}
\caption{Delaunay triangulation on the standard shape. The 15 landmarks (red points) form a mesh of 16 triangles. The central axis (L1, L9--L11, L2) divides the lung silhouette, while symmetric pairs define the bilateral contour.}
\label{fig:triangulacion}
\end{figure}

On the 15 landmarks of the standard shape, a Delaunay triangulation~\cite{delaunay1934sphere} was calculated, resulting in a mesh of 16 triangles covering the lung region (Figure~\ref{fig:triangulacion}). This mesh defines the correspondence between points of the original image and the standard shape.

For each pair of corresponding triangles (original image $\leftrightarrow$ standard shape), a unique affine transformation~\cite{wolberg1990digital} was calculated that maps the image within each triangle to its corresponding region in the standard shape. The transformation preserves continuity at triangle edges and uses bilinear interpolation.

\begin{figure}[!htbp]
\centering
\includegraphics[width=0.9\textwidth]{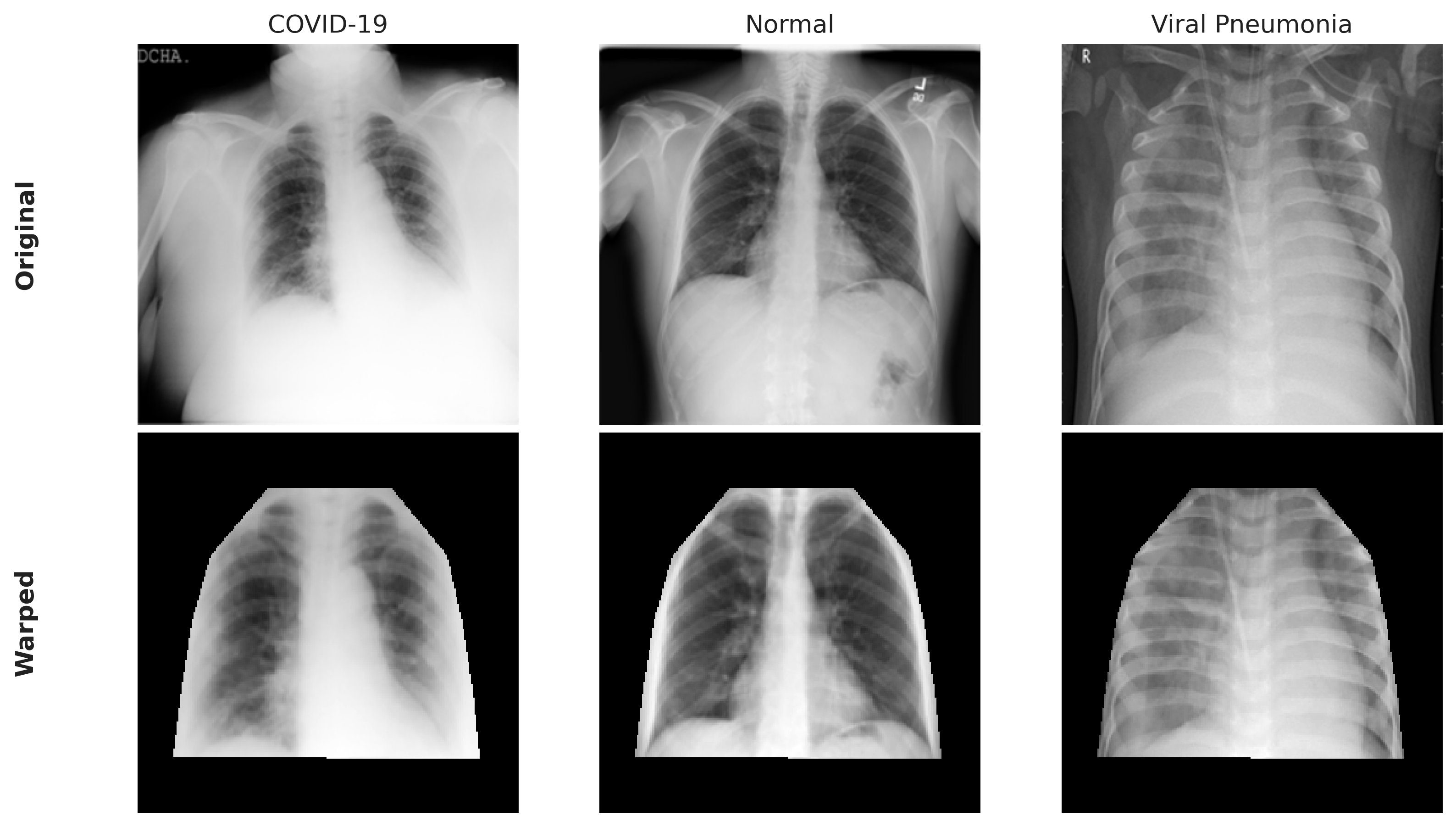}
\caption{Geometric normalization examples. Top: original radiographs showing variations in size, position, and orientation. Bottom: same images after warping to the standard shape.}
\label{fig:warping}
\end{figure}

A key parameter, \texttt{margin\_scale} = 1.05, controls a slight radial expansion from the landmark centroid, ensuring relevant peripheral anatomical context is included. The resulting normalized images have the lung region occupying approximately 47\% of the total area (Figure~\ref{fig:warping}). The complete dataset of 15,153 normalized images was generated using the ensemble model ($NN$ $1$) for landmark inference, constituting the input for training the classifier ($NN$ $2$).

\subsection{Classification}

For multiclass classification, a ResNet-18 architecture pretrained on ImageNet was used, replacing its final layer with a fully connected layer with 3 outputs (COVID-19, Normal, Viral Pneumonia). Normalized images were processed with \textbf{SAHS (Statistical Asymmetrical Histogram Stretching)}~\cite{CruzOvando2025SAHS}, a contrast enhancement method specifically designed for radiographs.

The dataset presents significant class imbalance (67\% Normal, 24\% COVID-19, 9\% Viral Pneumonia). To mitigate bias, weights inversely proportional to each class frequency were used in the Cross-Entropy loss function. Training employed data augmentation (horizontal flip, $\pm$10$^{\circ}$ rotations, slight affine transformations), AdamW optimizer~\cite{loshchilov2019adamw} with initial rate $1 \times 10^{-4}$, and early stopping based on validation set F1-Macro.

\subsection{Evaluation Metrics}

Performance was evaluated using: \textbf{Landmark error} (average Euclidean distance in pixels between predicted and annotated landmarks), \textbf{Accuracy}, \textbf{F1-Macro} (arithmetic mean of per-class F1-Score), and \textbf{Precision/Recall} per category.

\section{Results}

\subsection{Landmark Localization Precision}

The ensemble of four models with TTA achieved a mean error of \textbf{3.61 pixels} on 224$\times$224 images, representing a 10.6\% improvement over the best individual model (4.04 pixels). This error equals 1.6\% of image side size, sufficient precision for geometric normalization without introducing significant distortions.

\begin{table}[H]
\caption{Landmark detection precision. Error is measured in pixels on 224$\times$224 images.\label{tab:landmarks}}
\centering
\begin{tabularx}{\textwidth}{XCCC}
\toprule
\textbf{Configuration} & \textbf{Mean Error} & \textbf{Median Error} & \textbf{Improvement}\\
\midrule
Best individual model & 4.04 px & --- & ---\\
Ensemble (4 models) + TTA & 3.61 px & 3.07 px & 10.6\%\\
\bottomrule
\end{tabularx}
\end{table}

\begin{figure}[!htbp]
\centering
\includegraphics[width=0.95\textwidth]{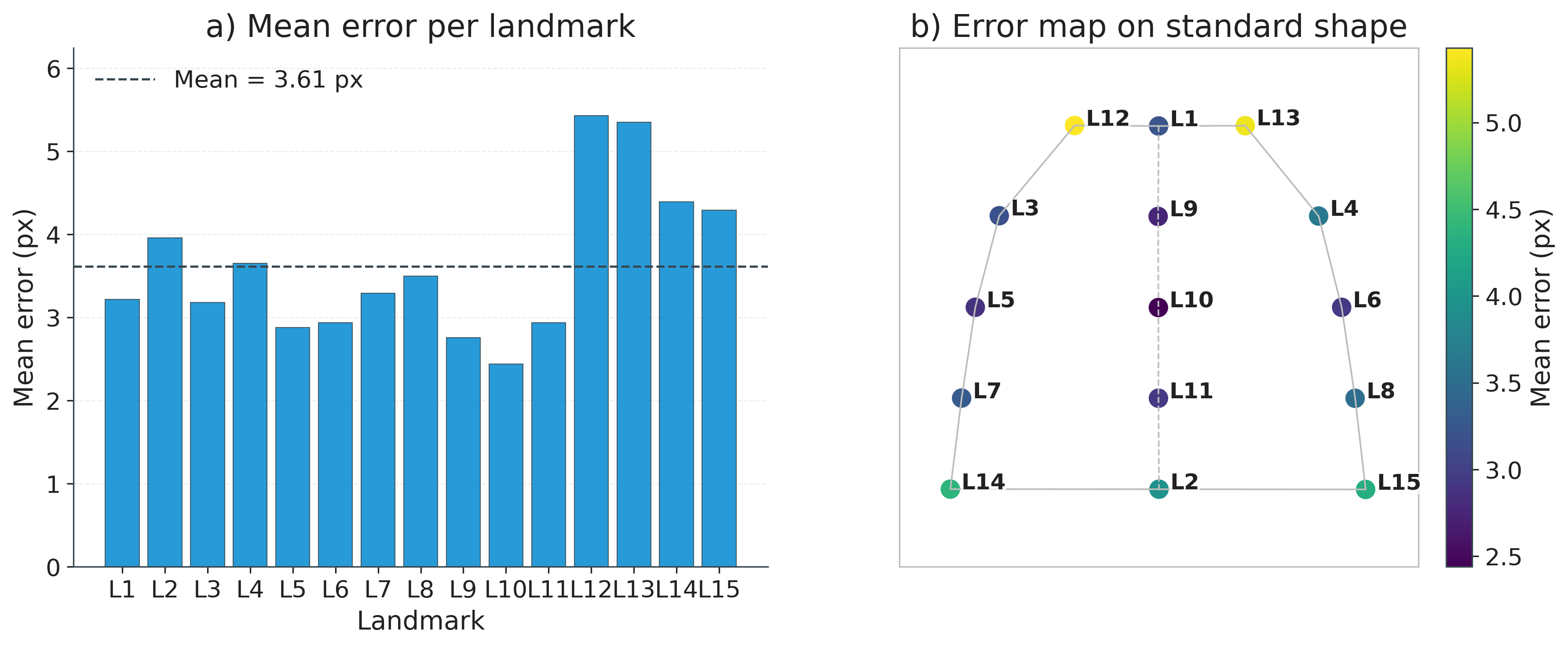}
\caption{Prediction error by landmark. (a) Mean error for each point L1--L15; dashed line indicates global mean (3.61 px). (b) Heat map on standard shape showing error distribution (2.5--5.0 px scale).}
\label{fig:error_landmarks}
\end{figure}

Per-landmark analysis revealed a systematic pattern (Figure~\ref{fig:error_landmarks}): higher precision at central axis points (2.44--2.94 pixels) due to clear definition of the vertebral midline, and lower precision at upper corners (5.35--5.43 pixels) where lung boundaries are less sharp. By category, error was consistent across pathologies: Normal (3.22 px), COVID-19 (3.93 px), Viral Pneumonia (4.11 px), demonstrating model robustness and lack of systematic bias.

\subsection{Effect of Normalization on Classification}

The classifier trained on normalized images achieved the following results on the five-fold cross-validation on the combined training and validation sets of 13,258 images: \textbf{Accuracy:} 98.60\% $\pm$ 0.26\%, \textbf{F1-Macro:} 98.00\% $\pm$ 0.36\%. The low standard deviations (all below 0.4 pp) confirm robustness and minimal dependence on specific training examples. Table~\ref{tab:rendimiento_clase} presents performance by category.

\begin{table}[H]
\caption{Classification average results by category (5-fold cross-validation).\label{tab:rendimiento_clase}}
\centering
\begin{tabularx}{\textwidth}{XCCCC}
\toprule
\textbf{Category} & \textbf{Precision} & \textbf{Recall} & \textbf{F1-score} & \textbf{Samples} \\
\midrule
COVID-19  & 98.08\% $\pm{0.32}$ & 98.23\% $\pm{0.59}$ & 98.15\% $\pm{0.36}$ & 3,164 \\
Viral Pneumonia & 96.61\% $\pm{2.13}$ & 97.19\% $\pm{1.19}$ & 96.87\% $\pm{0.79}$ & 1,176 \\
Normal    & \textbf{99.06\%} $\pm{0.08}$ & \textbf{98.91\%} $\pm{0.34}$ & \textbf{98.98\%} $\pm{0.20}$ & 8,918 \\
\bottomrule
\end{tabularx}
\end{table}

A controlled comparison between three preprocessing configurations (Table~\ref{tab:comparacion}) shows that the warped configuration achieves performance comparable to original images ($-$0.66 pp), while simple border cropping produces inferior results compared with warped images ($-$2.36 pp), demonstrating that geometric normalization effectively compensates for peripheral information loss.

\begin{table}[H]
\caption{5-fold cross-validation results for overall classifier performance. Note that the high accuracy in the first line should be interpreted with caution because peripheral artifacts may reveal class-related information. Therefore, that result cannot be considered valid.\label{tab:comparacion}}
\centering
\begin{tabularx}{\textwidth}{XCC}
\toprule
\textbf{Configuration} & \textbf{Accuracy (\%)} & \textbf{F1-macro (\%)} \\
\midrule
Original + SAHS & $99.26 \pm 0.18$ & $98.90 \pm 0.24$ \\
Warped + SAHS   & $98.60 \pm 0.26$ & $98.00 \pm 0.36$\\
Cropped + SAHS & $96.24 \pm 0.29$ & $95.36 \pm 0.32$ \\
\bottomrule
\end{tabularx}
\end{table}

\begin{figure}[!htbp]
\centering
\includegraphics[width=0.6\textwidth]{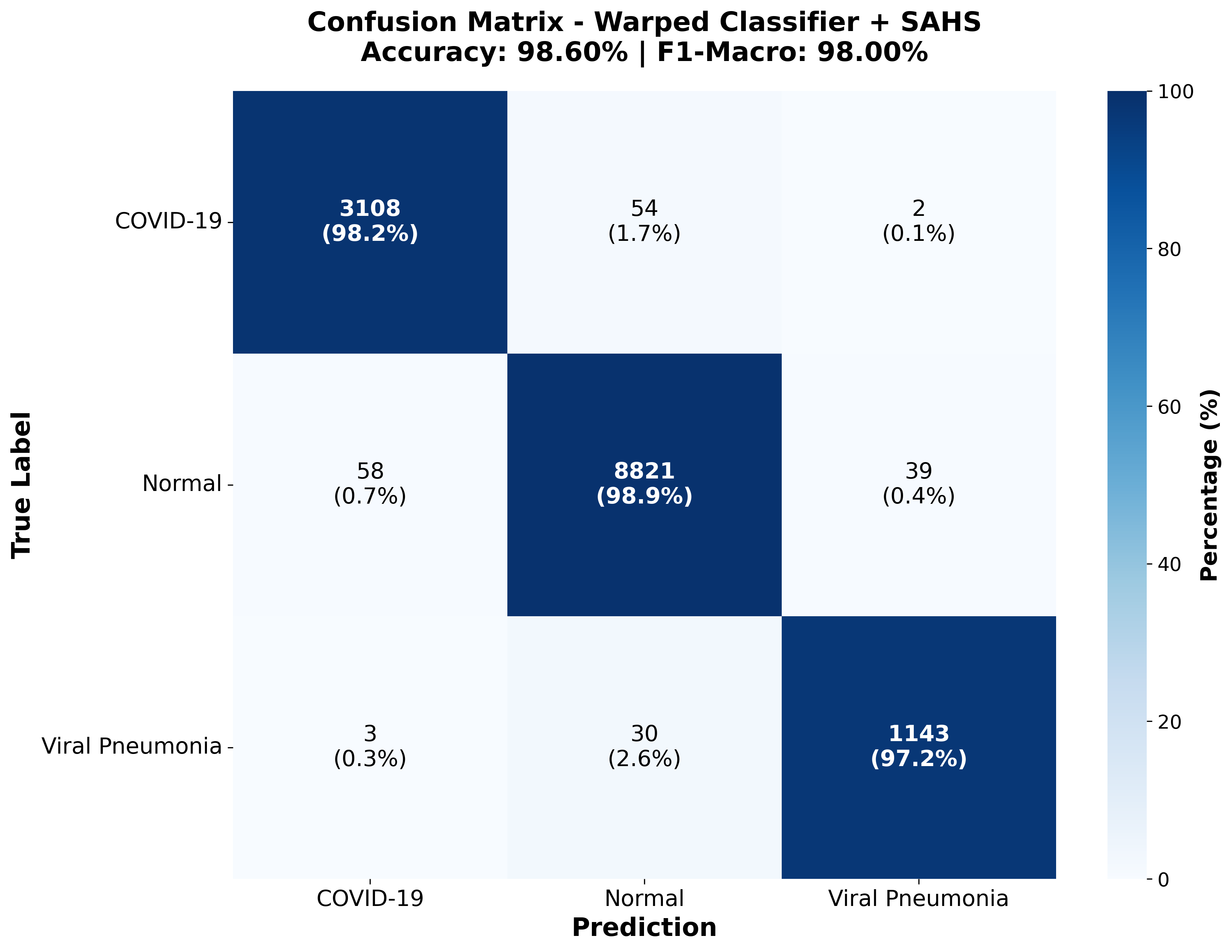}
\caption{Confusion matrix of the Warped + SAHS classifier obtained from the aggregated out-of-fold predictions of the five-fold cross-validation (13,258 samples; Accuracy: 98.60\%, F1-Macro: 98.00\%). Confusion between COVID-19 and Viral Pneumonia is minimal (5 cases).}
\label{fig:matriz_confusion}
\end{figure}

The confusion matrix associated with the best-performing results corresponds to the dataset with the largest number of samples (13,258 out-of-fold predictions from the five-fold cross-validation) (Figure~\ref{fig:matriz_confusion}) and reveals that the main confusions occur between Normal and COVID-19 (112 cases) and between Normal and Viral Pneumonia (69 cases). Confusion between COVID-19 and Viral Pneumonia is minimal (only 5 of 13,258 cases, 0.04\%), which is clinically relevant since these conditions require different treatments.

\begin{figure}[!htbp]
\centering
\includegraphics[width=0.95\textwidth]{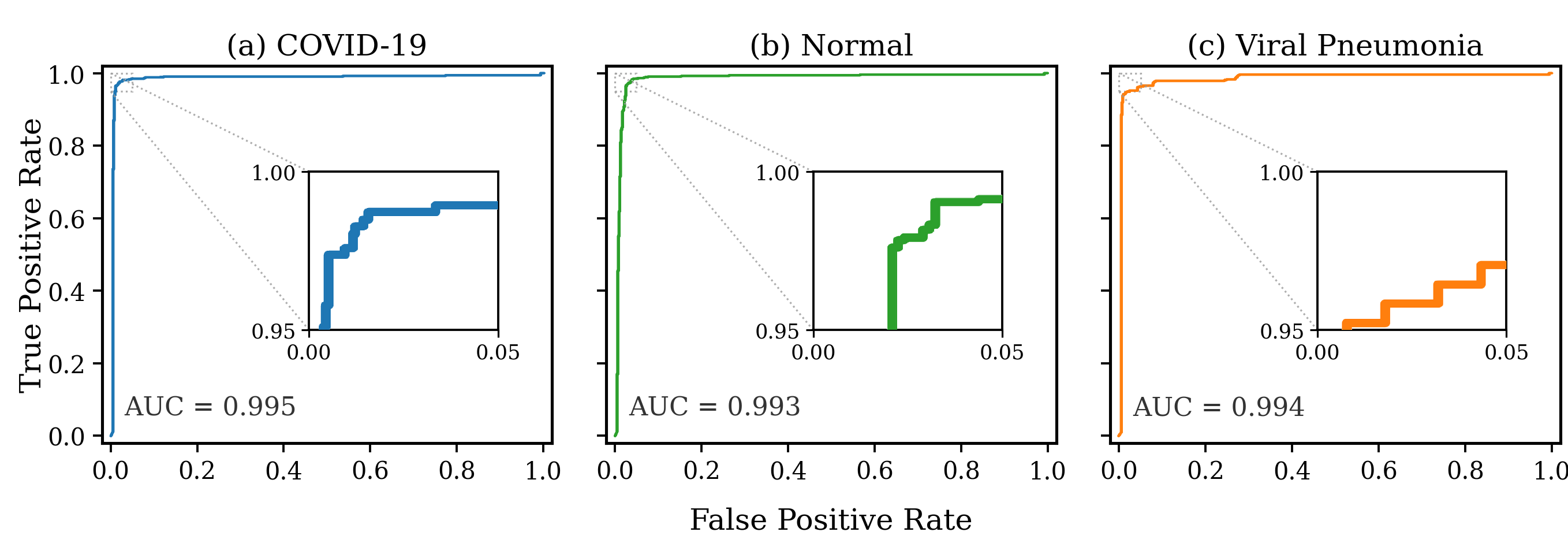}
\caption{ROC curves (one-vs-rest): (a) COVID-19 vs. rest (AUC=0.995), (b) Normal vs. rest (AUC=0.993), (c) Viral Pneumonia vs. rest (AUC=0.994). Insets zoom into the low false positive rate region.}
\label{fig:roc_curves}
\end{figure}

ROC curves (Figure~\ref{fig:roc_curves}) confirm excellent discrimination for all three classes, with AUC values exceeding 0.99.

Additional experiments were conducted using a second dataset to evaluate the impact of normalization on classification performance. Retraining and testing procedures were carried out to validate the relevance and effectiveness of image normalization. The specific characteristics of this second dataset made it possible to strengthen the evaluation of the benefits associated with normalization techniques by incorporating more diverse samples, including adult and pediatric patients.

The second dataset consists of 9,000 images selected to incorporate characteristics from different age ranges, aiming to increase the generalization capability of the neural network employed. A subset of the Kermany dataset~\cite{kermany2018identifying}, specifically focused on pediatric cases, was used for this purpose. Thus, the new training dataset, using data normalization, is balanced with the following distribution: 3,000 instances of COVID-19, 3,000 instances of pneumonia, of which 1,500 correspond to pediatric cases, finally 3,000 normal cases, with 1,500 adult samples and an equal number of pediatric samples.

Thus, the class distribution was organized as follows: the COVID-19 class was composed exclusively of adult cases due to the insufficient number of pediatric radiographs available; the pneumonia class consisted of $50\%$ adult cases and $50\%$ pediatric cases from the Kermany dataset; similarly, the normal class included $50\%$ adult cases and $50\%$ pediatric cases.

Similarly, to incorporate relevant information despite not having a specific label distinction within the pneumonia category in the target dataset, the labeling scheme from the Kermany dataset was leveraged, as it differentiates between viral and bacterial pneumonia cases. Therefore, the pediatric subset was divided such that, among the 1,500 pediatric instances included, 750 corresponded to viral pneumonia cases and 750 to bacterial pneumonia cases.

\begin{figure}[!htbp]
\centering
\includegraphics[width=0.5\textwidth]{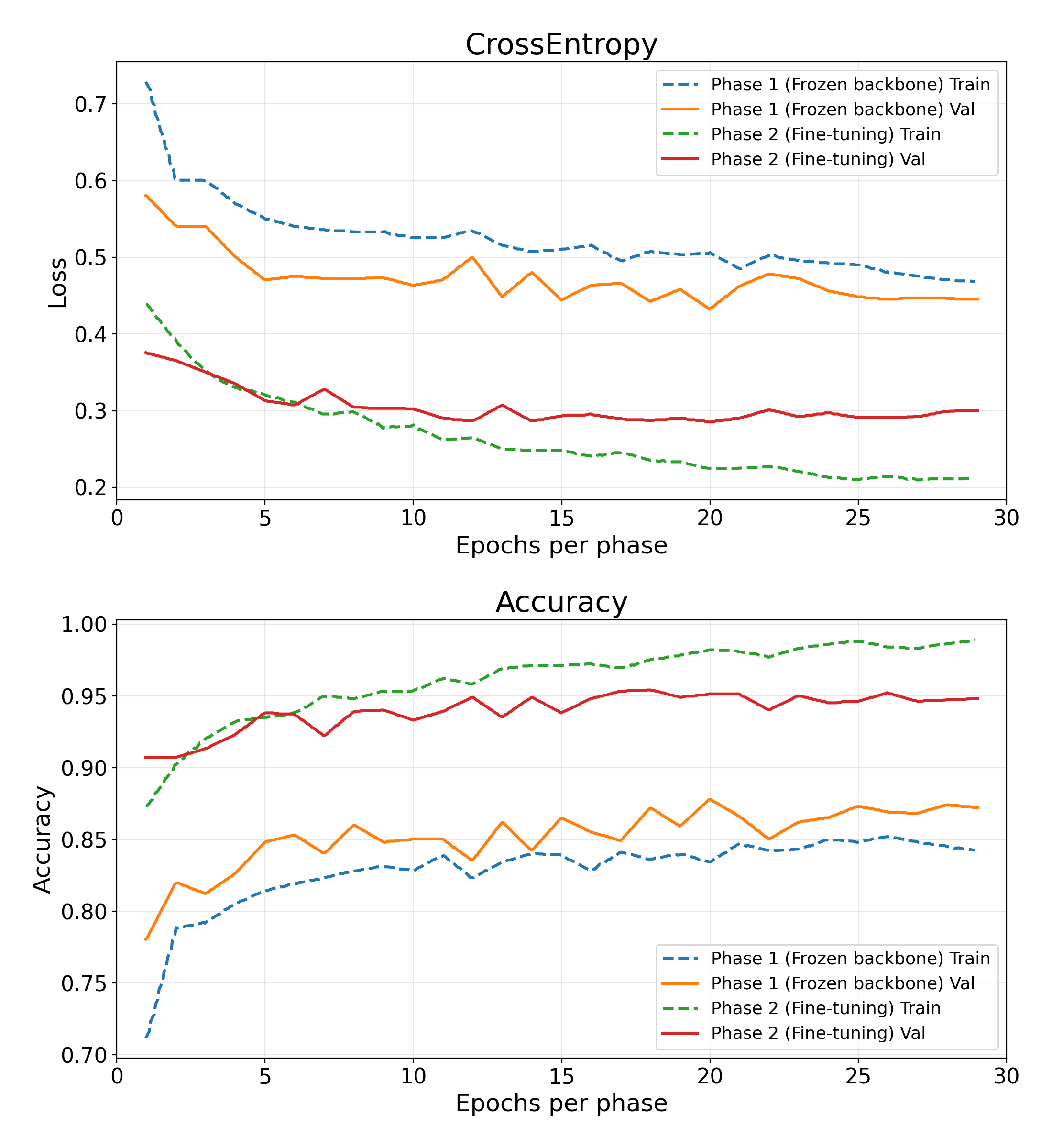}
\caption{Model learning curves, including the evolution of the loss function and accuracy during the training and validation phases.}
\label{fig:curva_entrenamiento}
\end{figure}

Figure~\ref{fig:curva_entrenamiento} shows the learning curves corresponding to the training process performed (70\% dataset) using the balanced dataset, which incorporates information from both adult and pediatric populations.

\begin{figure}[!htbp]
\centering
\includegraphics[width=0.6\textwidth]{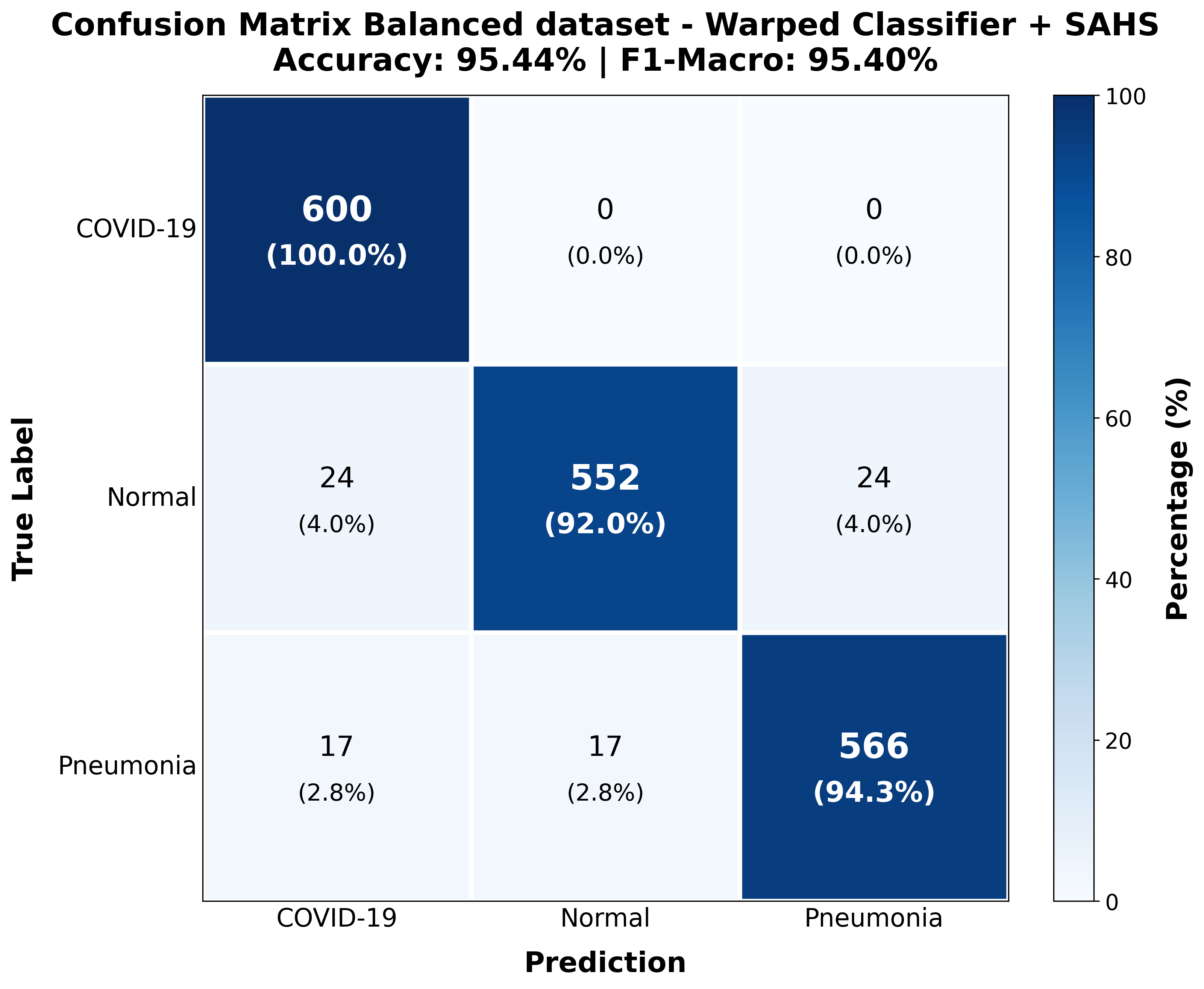}
\caption{Confusion matrix of the first repeated split of the Warped + SAHS balanced validation dataset (Accuracy: 95.44\%, F1-Macro: 95.40\%).}
\label{fig:matriz_confusion0}
\end{figure}

\begin{figure}[!htbp]
\centering
\includegraphics[width=0.6\textwidth]{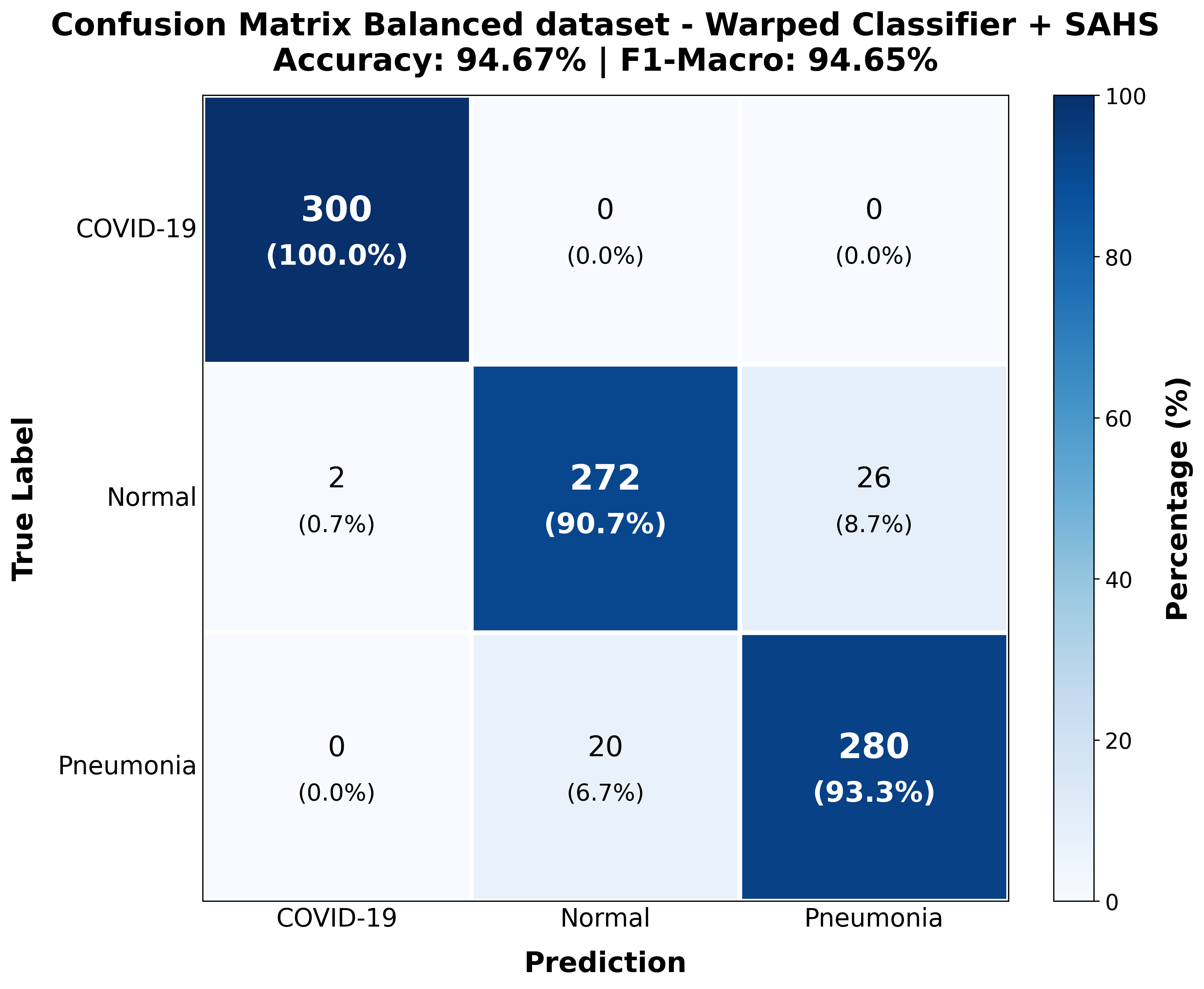}
\caption{Confusion matrix of the first repeated split of the Warped + SAHS balanced test dataset (Accuracy: 94.67\%, F1-Macro: 94.65\%).}
\label{fig:matriz_confusion1}
\end{figure}

Table~\ref{tab:metricas_clasificacion1} and Table~\ref{tab:metricas_clasificacion2} present the respective averaged classification metrics for the balanced dataset after retraining the ResNet18 model. These values correspond to the average metrics across the five repeated stratified train/validation/test splits, whereas Figures~\ref{fig:matriz_confusion0} and \ref{fig:matriz_confusion1} show the confusion matrices obtained from the first repeated split as representative examples.

\begin{table}[H]
\centering
\caption{Classification average results of the retrained model on the validation set ($20\%$ of the data).}
\label{tab:metricas_clasificacion1}
\begin{tabular}{lcccc}
\hline
\textbf{Category} & \textbf{Precision} & \textbf{Recall} & \textbf{F1-score} & \textbf{Samples} \\
\hline
COVID-19     & 99.67\% $\pm{0.15}$ & \textbf{99.50\%} $\pm{0.35}$ & 99.58\%   $\pm{0.23}$ & 600 \\
Pneumonia  & \textbf{94.52\%} $\pm{0.75}$ & 92.00\%   $\pm{0.69}$ & 93.24\% $\pm{0.51}$ & 600 \\
Normal    & 92.18\% $\pm{1.16}$ & \textbf{94.33\%}     $\pm{0.98}$ & 93.24\% $\pm{0.75}$ & 600 \\
\hline
\end{tabular}
\end{table}

\begin{table}[H]
\centering
\caption{Classification average results of the retrained model on the test set ($10\%$ of the data).}
\label{tab:metricas_clasificacion2}
\begin{tabular}{lcccc}
\hline
\textbf{Category} & \textbf{Precision} & \textbf{Recall} & \textbf{F1-score} & \textbf{Samples} \\
\hline
COVID-19     & 99.34\% $\pm{0.52}$& \textbf{99.50\%} $\pm{0.5}$& 99.42\% $\pm{0.36}$ & 300 \\
Pneumonia  & \textbf{93.15\%} $\pm{0.89}$ & 90.67\% $\pm{1.36}$& 91.89\% $\pm{0.82}$& 300 \\
Normal    & 91.50\% $\pm{1.09}$& \textbf{93.33\%} $\pm{0.95}$& 92.41\% $\pm{0.72}$& 300 \\
\hline
\end{tabular}
\end{table}

Table~\ref{tab:comparacionBalanceo} presents the experiments conducted using the previously described balanced dataset. It can be observed that the best performance is achieved when using chest X-ray images without cropping or alignment. This behavior may be attributed to the presence of well-defined artifacts along the image borders, which differ across classes and facilitate their discrimination. When cropping or alignment procedures are applied, these artifacts, which can be consistently identified for each class, are removed. As a result, the detection rate decreases; however, confidence increases that the model is making its decisions based on patterns associated with pulmonary pathologies rather than on spurious features or imaging artifacts.

\begin{table}[H]
\caption{Five repeated stratified train/validation/test splits results for overall classifier performance. Note that the accuracy in the first line is very high due to the presence of artifacts in the image that reveal the class. Therefore, that result cannot be considered valid.\label{tab:comparacionBalanceo}}
\centering
\begin{tabularx}{\textwidth}{XCC}
\toprule
\textbf{Configuration} & \textbf{Accuracy} & \textbf{F1-Macro}\\
\midrule
Original + SAHS & 97.55\% $\pm{1.02}$& 97.11\% $\pm{0.97}$\\
Warped + SAHS (proposed Balanced dataset)& {94.67\%} $\pm{0.95}$& {94.66\%} $\pm{0.65}$\\
Cropped (12\%) + SAHS & 94.17\% $\pm{0.79}$& 93.88\% $\pm{0.76}$\\
\bottomrule
\end{tabularx}
\end{table}

The confusion matrices corresponding to the balanced dataset for the validation and test subsets are presented in Figures~\ref{fig:matriz_confusion0} and \ref{fig:matriz_confusion1}, respectively. As shown in Figure~\ref{fig:matriz_confusion0}, which corresponds to the validation confusion matrix, the main area for improvement lies in the misclassification of healthy patients (normal) as pneumonia or COVID-19 cases. To a lesser extent, pneumonia cases are confused with COVID-19 or classified as healthy patients.

Conversely, Figure~\ref{fig:matriz_confusion1}, which corresponds to the test subset, reveals that the predominant source of misclassification stems from healthy patients being predicted as pneumonia cases, whereas, to a lesser degree, pneumonia cases are misclassified as healthy patients.

Because COVID-19 cases were exclusively adult while the pneumonia and normal classes included both adult and pediatric images, this balanced dataset may still contain residual age- and source-related biases. Therefore, these results should be interpreted as an additional robustness experiment rather than as definitive external validation.

\subsection{Model Interpretability Through Attention Visualization}

\begin{figure}[!htbp]
\centering
\includegraphics[width=0.8\textwidth]{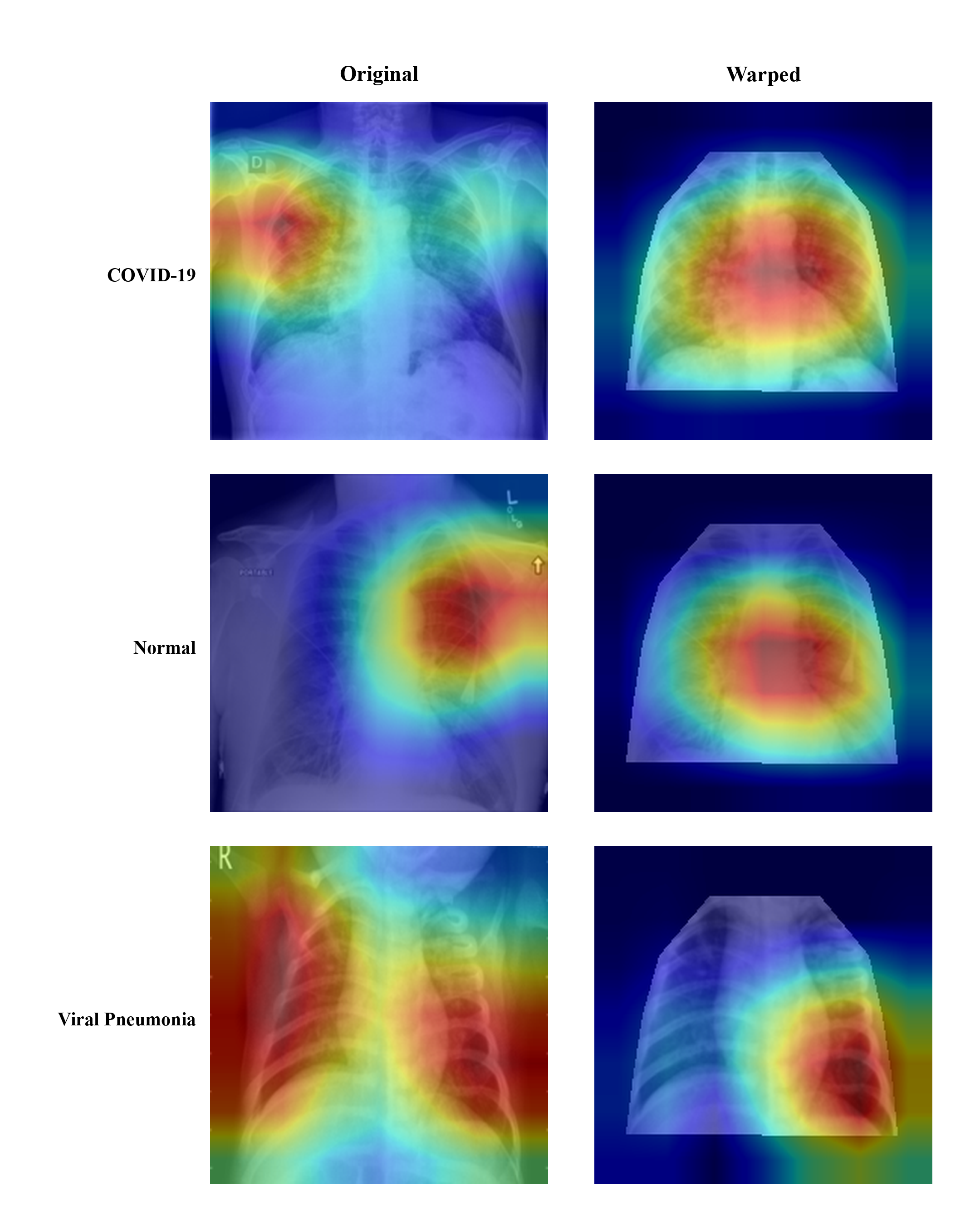}
\caption{Grad-CAM visualization comparing original (left) and warped (right) images for COVID-19, Normal, and Viral Pneumonia cases. Warped images show attention concentrated within lung regions rather than on peripheral artifacts.}
\label{fig:gradcam}
\end{figure}

Grad-CAM visualization (Figure~\ref{fig:gradcam}) confirms that attention in original images often concentrates on artifacts, while attention in warped images remains within lung regions, supporting the hypothesis that geometric normalization constrains the model to learn clinically relevant features.

\subsection{Failure Case Analysis}

\begin{figure}[!htbp]
\centering
\includegraphics[width=0.95\textwidth]{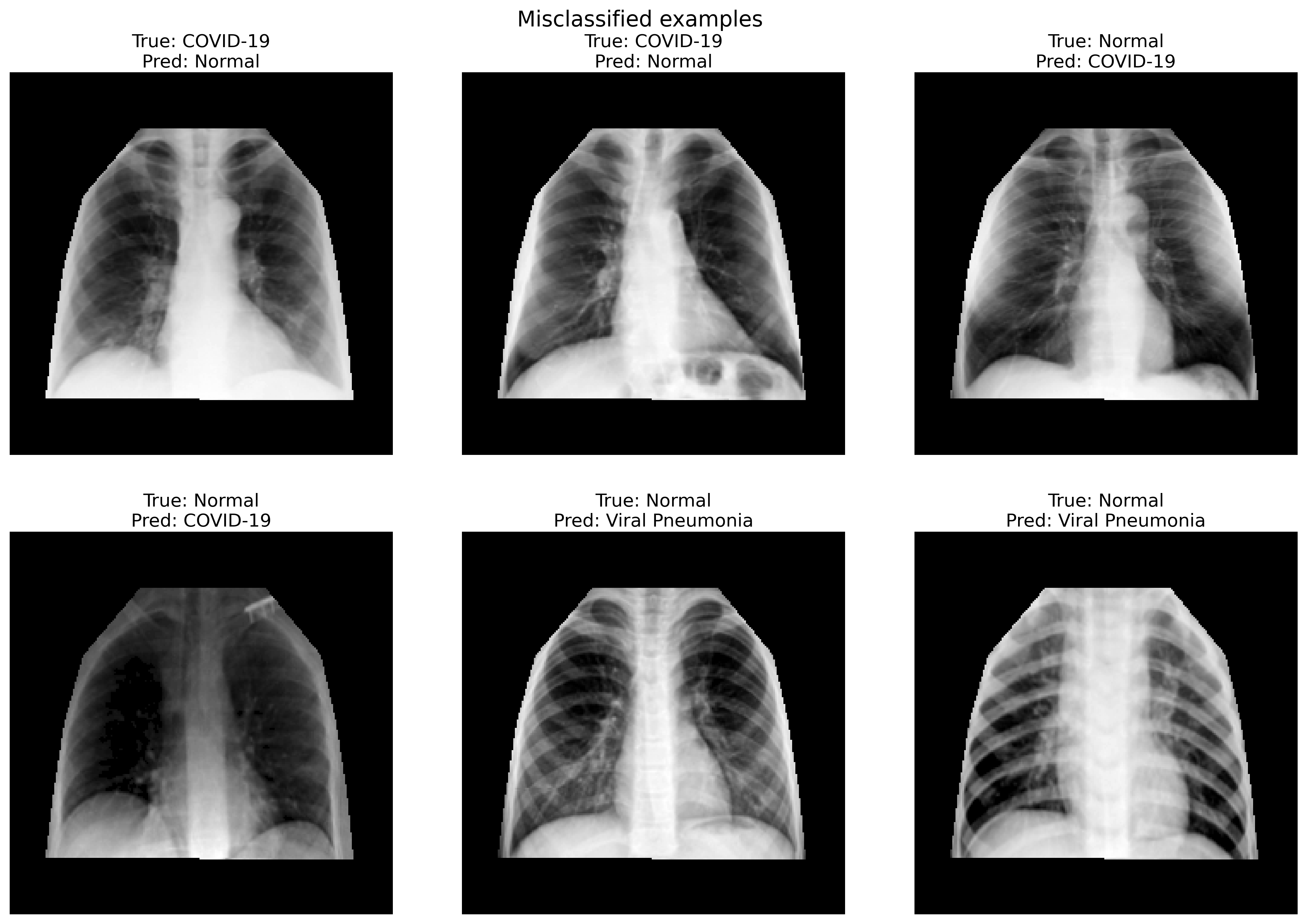}
\caption{Representative failure cases showing characteristic misclassification patterns. Top: COVID-19 cases with subtle opacities misclassified as Normal, and a Normal case with prominent vascular markings misclassified as COVID-19. Bottom: Normal cases misclassified as pathological due to artifacts or prominent anatomical structures.}
\label{fig:failure_cases}
\end{figure}

The best results obtained from the conducted experiments demonstrate that the analysis of the 186 misclassified cases (1.40\% error rate) from the five-fold cross-validation revealed characteristic patterns (Figure~\ref{fig:failure_cases}, Table~\ref{tab:error_patterns}). Manual review confirmed that 24/186 (12.9\%) of the misclassified cases presented ambiguous visual features where inter-expert disagreement would be expected.

\begin{table}[H]
\caption{Distribution of misclassification patterns in the five-fold cross-validation (186 total errors from 13,258 cases).\label{tab:error_patterns}}
\centering
\begin{tabularx}{\textwidth}{XCCC}
\toprule
\textbf{Error Type} & \textbf{Count} & \textbf{\% of Class} & \textbf{Likely Cause}\\
\midrule
COVID-19 $\rightarrow$ Normal & 54 & 1.71\% & Subtle opacities, early/late stage\\
Normal $\rightarrow$ COVID-19 & 58 & 0.65\% & Prominent vessels, soft tissue overlap\\
Viral Pneumonia $\rightarrow$ Normal & 30 & 2.55\% & Minimal infiltrates, atypical presentation\\
Normal $\rightarrow$ Viral Pneumonia & 39 & 0.44\% & Peribronchial markings, pediatric pattern\\
Viral Pneumonia $\rightarrow$ COVID-19 & 3 & 0.26\% & Diffuse opacities shared between pneumonias\\
\textbf{COVID-19 $\rightarrow$ Viral Pneumonia} & \textbf{2} & \textbf{0.06\%} & \textbf{Diffuse opacities shared between pneumonias}\\
\bottomrule
\end{tabularx}
\end{table}

Two key observations emerge: (1) confusion between COVID-19 and Viral Pneumonia is minimal (5 cases, 0.04\% of all samples), which is clinically significant as these conditions require different treatment protocols; and (2) Viral Pneumonia exhibits the highest error rate (2.81\%), attributable to class imbalance and greater presentation heterogeneity. False positives are rare, only 97 Normal cases (1.09\%) were misclassified as pathological.

\subsection{Comparison with Literature}

Table~\ref{tab:comparacion_literatura} compares the proposed system with related works. The system achieves competitive performance using a larger dataset (15,153 images), classifying 3 classes, and employing a lighter architecture (ResNet-18), with the added advantage of interpretability through explicit landmark visualization and transformation auditability.

\begin{table}[H]
\caption{Comparison with related works in COVID-19 classification.\label{tab:comparacion_literatura}}
\centering
\begin{tabularx}{\textwidth}{XCCCC}
\toprule
\textbf{Work} & \textbf{Classes} & \textbf{Dataset} & \textbf{Architecture} & \textbf{Accuracy}\\
\midrule
\textbf{Proposed system} & 3 & 15,153 & ResNet-18 + Warping & \textbf{98.60\%}\\
\midrule
Chowdhury et al.~\cite{chowdhury2020can} & 4 & 423 & VGG19 & 96.58\%\\
Rahman et al.~\cite{rahman2021exploring} & 3 & 3,616 & VGG16 & 93.94\%\\
Ozturk et al.~\cite{ozturk2020automated} & 2 & 1,127 & DarkCovidNet & 98.08\%\\
Narin et al.~\cite{narin2021automatic} & 2 & 341 & ResNet-50 & 98.00\%\\
\bottomrule
\end{tabularx}
\end{table}

\section{Discussion}

\subsection{Interpretation of Landmark Detection Results}

The ensemble achieved a mean error of 3.61 pixels (1.14\% of image diagonal). Assuming a typical posteroanterior chest radiograph field of view of 32 cm, this translates to approximately 5 mm in physical space, this precision is comparable to automated lung field detection systems reported in the literature (5--10 pixels) and superior to many intensity-based segmentation approaches. The systematic error pattern reflects the underlying anatomy: central axis landmarks (L9--L11) achieve highest precision (2.44--2.94 pixels) due to the clear reference provided by the vertebral column, which appears as a consistent high-intensity vertical structure in chest radiographs, while upper corner landmarks (L12, L13) exhibit the largest errors (5.35--5.43 pixels) where lung boundaries are inherently diffuse due to superposition with mediastinal structures. A possible explanation for the higher errors observed at landmarks L12 and L13 is the greater variability of the local anatomical neighborhood around these points. In the available training images, the anatomical structures surrounding these landmarks may differ substantially between pediatric and adult radiographs, especially because bone size, thoracic proportions, and upper-lung boundary appearance can change considerably across these two groups. In contrast, landmarks located near more stable anatomical regions, such as L10, tend to exhibit a more consistent local structure across pediatric and adult cases, which may explain their lower localization error. Future work could address this variability by using age-stratified training sets, landmark-specific loss weights, or multi-scale feature extraction modules that emphasize anatomical cues in highly variable boundary regions.

The consistency across disease categories (Normal: 3.22 px, COVID-19: 3.93 px, Viral Pneumonia: 4.11 px) confirms robustness without systematic bias, with the modest increase in pathological cases attributable to boundary distortion from infiltrates and consolidations.

\subsection{Effect of Geometric Normalization on Classification Performance}

The 0.66 pp difference between Original and Warped configurations, while numerically small, is conceptually relevant. Original images may contain discriminative information outside the lung region, for example, institutional labels and acquisition artifacts that correlate with disease severity. This hypothesis is supported by Zech et al.~\cite{zech2018variable}, who demonstrated that pneumonia detection models could predict the originating hospital with 99.95\% accuracy, exploiting spurious correlations.

The cropped configuration (96.24\%) confirms that peripheral borders contribute discriminative power to the original model. The warped configuration (98.60\%) outperforms cropping by 2.36 pp, demonstrating that geometric normalization compensates for information loss through spatial standardization. Thus, the 98.60\% accuracy of the warped model provides a more conservative and clinically relevant estimate based exclusively on lung features.

Conceptually, geometric normalization acts as a spatial filter with selective preservation properties: it retains local texture (ground-glass opacities, consolidations, interstitial infiltrates) and intensity patterns within the lung while eliminating absolute spatial position, global scale, orientation, and extra-pulmonary information (markers, labels). This selective filtering encourages the classifier to learn representations more closely related to pulmonary pathology and reduces the risk of relying on extra-pulmonary artifacts.

\subsection{Comparison with Alternative Normalization Approaches}

Spatial Transformer Networks (STN)~\cite{jaderberg2015spatial} represent a prominent end-to-end alternative for geometric normalization. However, STNs present critical limitations for medical imaging: they apply a single global affine transformation that is inadequate for deformable organs exhibiting non-rigid variability, require substantial training data as they lack anatomical priors, and produce non-auditable transformations that hinder clinical validation. In contrast, our piecewise affine warping applies 16 local transformations guided by explicit anatomical landmarks, enabling non-uniform deformations that capture regional lung shape variability. The modular approach sacrifices joint optimization but gains interpretability (landmarks are visually verifiable), lower data requirements (957 manually annotated examples), and reusability (the landmark detector can serve other tasks).

\subsection{Clinical Implications}

The complete inference pipeline executes in under 1 second on GPU ($\sim$3 seconds on CPU), compatible with real-time applications including second opinion for general radiologists and mass screening in high-throughput settings. A critical advantage is multi-level interpretability: unlike black-box classifiers, the system decomposes the classification process into auditable stages: landmark localization (verifiable by radiologists), geometric transformation (inspectable through side-by-side visualization), and classification (complemented with Grad-CAM attention maps). This transparency could allow clinicians to identify potential failures, for instance if landmarks fall outside lung boundaries or if the warped image exhibits severe distortions, and reject unreliable predictions, addressing a fundamental barrier to clinical adoption of deep learning systems.

It is important to emphasize that the proposed system is intended exclusively as a machine learning research prototype algorithm, not as a standalone diagnostic system. Its predictions should be interpreted by qualified healthcare professionals in conjunction with clinical findings, radiological assessments, laboratory tests, and institutional diagnostic protocols.

\subsection{Limitations and Future Directions}

This study has several limitations:

\begin{itemize}
\item An important limitation of the present study is the absence of a true external validation on independent multi-institutional datasets. Although the balanced adult--pediatric mixed dataset provides an additional robustness experiment, it should not be interpreted as definitive external validation. Future work should evaluate the proposed pipeline on independent datasets such as MIMIC-CXR, CheXpert, PadChest, or other multi-center radiographic collections to assess generalization across acquisition protocols, patient populations, and imaging devices.

\item Only ResNet-18 was evaluated as a classifier; alternative architectures (DenseNet-121, EfficientNet-B0, Vision Transformers) may benefit differently from geometric normalization.

\item The landmark annotation of 957 images, while substantially less than pixel-level segmentation, represents a non-trivial burden that could be reduced through semi-supervised or few-shot learning strategies.
\end{itemize}

Future work could redefine them to align with standardized radiology terminology (Fleischner Society, ACR). For pathologies where geometric variability is diagnostically informative (e.g., hyperinflation in COPD, pneumothorax), a hybrid approach extracting geometric features alongside normalized texture may be needed.

Future directions include: (1) extension to additional pathologies (tuberculosis, pulmonary edema, nodules), (2) comparison with end-to-end STN-based methods to quantify the interpretability-performance trade-off, and (3) exploitation of the fixed anatomical layout of normalized images for lesion localization: because the alignment procedure maps each anatomical landmark to a consistent spatial position, it may be possible to generate standardized lesion maps that indicate not only the presence of a pathology but also its precise anatomical location, potentially improving the interpretability and clinical utility of the system.

\section{Conclusions}

This work presented a pulmonary disease classification system based on anatomical landmark detection and geometric normalization of the lung region. The results support the following conclusions:

\begin{itemize}
\item Geometric alignment provides a controlled representation of chest X-rays by reducing variations in position, scale, orientation, and extra-pulmonary artifacts.
\item Original images achieved slightly higher raw accuracy; however, Grad-CAM analysis showed attention on artifacts, indicating that part of this advantage may come from non-pathological cues.
\item Although original images achieved the highest raw accuracy, this result is likely influenced by peripheral acquisition artifacts. When compared against artifact-masked unaligned images, the geometrically normalized representation achieved higher performance: (98.60\% vs.\ 96.24\%) for the COVID-19 Radiography Database and (94.67\% vs.\ 94.17\%) for the balanced adult--pediatric mixed dataset including pediatric cases from the Kermany dataset, supporting the benefit of anatomical alignment under a more controlled evaluation setting. These results suggest that geometrically normalized images can improve classification performance under an artifact-controlled evaluation setting.
\item The classifier achieved stable, high performance under five-fold cross-validation (98.60$\pm$0.26\% accuracy, 98.00\% F1-Macro).
\item Geometric alignment ensures that anatomical regions are consistently mapped to fixed spatial positions across all images. As a result, the discriminative features relevant to each pathology always appear at the same location in the normalized image, reducing the representational burden on the classifier. Although convolutional neural networks can tolerate variations in translation, scale, and rotation through their inductive biases, such variability makes it harder to reliably abstract the most discriminative features. By eliminating this variability, geometric normalization allows the classifier to focus its capacity on pathology-specific patterns rather than on compensating for positional differences.
\end{itemize}

The main contributions of this work are:

\begin{itemize}
\item A robust automatic landmark detector for lung contour localization, achieving a mean error of 3.61 pixels through an ensemble with test-time augmentation.
\item A complete geometric normalization framework combining Generalized Procrustes Analysis, Delaunay triangulation, and piecewise affine warping.
\item An automated method for generating an image of the lung region, standardized in shape, pose and size, which could be useful in future work to obtain a precise anatomical mapping of lung lesions.
\item An interpretable pipeline in which landmark prediction, image warping, classification, and Grad-CAM visualization can be inspected independently.
\end{itemize}

These results provide methodological evidence on the importance of controlling extrinsic variability in medical images to develop classifiers that learn pathology-relevant patterns rather than spurious correlations---a critical property for future clinical validation and potential decision-support deployment across institutions.

\section*{Data Availability}
The original python code developed to produce all the results reported in this paper is openly available at \url{https://doi.org/10.5281/zenodo.18626394} and \url{https://doi.org/10.5281/zenodo.18780919}.

\begingroup
\small
\setlength{\bibsep}{2pt plus 0.3ex}
\bibliographystyle{unsrtnat}
\bibliography{references}
\endgroup

\end{document}